\documentclass[8pt,twocolumn]{IEEEtran}

\makeatletter
\def\ifundefined{\@ifundefined}
\makeatother

\usepackage{epsfig}
\usepackage{dsfont}
\usepackage{amsmath}
\usepackage{amssymb} 
\usepackage{graphicx}
\usepackage{amsthm}
\usepackage[normalem]{ulem}
\usepackage{subfigure}
\usepackage{color}
\usepackage{hyperref}
\usepackage{bm}
\usepackage{lineno}
\usepackage{multirow}
\usepackage[ruled,vlined]{algorithm2e}
\usepackage{amsmath}
\usepackage{cite}
\usepackage{lettrine}
\usepackage{diagbox} %For Mac
\usepackage{rotating}
\usepackage[super]{nth}

\DeclareMathOperator{\Ex}{\mathbb{E}}% expected value

\makeatletter
\let\start@align@nopar\start@align
\let\start@gather@nopar\start@gather
\let\start@multline@nopar\start@multline
\long\def\start@align{\par\start@align@nopar}
\long\def\start@gather{\par\start@gather@nopar}
\long\def\start@multline{\par\start@multline@nopar}
\makeatother
\newcommand{\be}{\begin{equation}}
\newcommand{\ee}{\end{equation}}
\newcommand{\bea}{\begin{eqnarray}}
\newcommand{\eea}{\end{eqnarray}}
\newcommand\ba[1]{\left[ \begin{array}{#1}}
	\def\ea{\end{array}\right]}

\newcommand{\bfi}{\begin{figure}}
	\newcommand{\efi}{\end{figure}}

\newcommand{\R}{\mathbb{R}}

\newcommand{\X}{\mathcal{X}}

\newcommand{\x}{\mathbf{x}}

\newcommand{\G}{\mathbf{G}}
\newcommand{\cmmnt}[1]{}

\newcommand{\inner}[1]{\langle {#1} \rangle}
\newcommand{\overbar}[1]{\mkern 1.5mu\overline{\mkern-1.5mu#1\mkern-1.5mu}\mkern 1.5mu}

\providecommand{\norm}[1]{\lVert#1\rVert}

\newtheorem{thm}{Theorem}  %   number by Chapter
\newtheorem{pro}{Proposition}

\newtheorem{defn}{Definition}       %   Numbered independently from thm
\begin{document}
	%\linenumbers
	\title{Fast Estimation of Information Theoretic Learning Descriptors using Explicit Inner Product Spaces
		\thanks{This work was supported by the Lifelong Learning Machines program from DARPA/MTO grant FA9453-18-1-0039.}
		\thanks{The authors are with the Computational NeuroEngineering Laboratory, University of Florida, Gainesville, FL 32611 USA  (e-mail: likan@ufl.edu; principe@cnel.ufl.edu).}}
	\author{Kan Li, \IEEEmembership{Member,~IEEE} and Jos\'{e} C. Pr\'{i}ncipe, \IEEEmembership{Fellow,~IEEE}}
	\markboth{}%
	{Li \MakeLowercase{\textit{et al.}}: }	
	\maketitle
	
\begin{abstract}
Kernel methods form a theoretically-grounded, powerful and versatile framework to solve nonlinear problems in signal processing and machine learning.	The standard approach relies on the \emph{kernel trick} to perform pairwise evaluations of a kernel function, leading to scalability issues for large datasets due to its linear and superlinear growth with respect to the training data. Recently, we proposed \emph{no-trick} (NT) kernel adaptive filtering (KAF) that leverages explicit feature space mappings using data-independent basis with constant complexity. The inner product defined by the feature mapping corresponds to a positive-definite finite-rank kernel that induces a finite-dimensional reproducing kernel Hilbert space (RKHS). Information theoretic learning (ITL) is a framework where information theory descriptors based on non-parametric estimator of R\'{e}nyi entropy replace conventional second-order statistics for the design of adaptive systems. An RKHS for ITL defined on a space of probability density functions simplifies statistical inference for supervised or unsupervised learning. ITL criteria take into account the higher-order statistical behavior of the systems and signals as desired. However, this comes at a cost of increased computational complexity. In this paper, we extend the NT kernel concept to ITL for improved information extraction from the signal without compromising scalability. Specifically, we focus on a family of fast, scalable, and accurate estimators for ITL using explicit inner product space (EIPS) kernels. We demonstrate the superior performance of EIPS-ITL estimators and combined NT-KAF using EIPS-ITL cost functions through experiments.
\end{abstract}
	
\section{Introduction}
Information theoretic learning (ITL) is a framework where information theory descriptors based on non-parametric estimator of R\'{e}nyi entropy replace conventional second-order statistics for the design of adaptive systems \cite{ITL}. A reproducing kernel Hilbert space (RKHS) for ITL defined on a space of probability density functions (pdf's) simplifies statistical inference for supervised or unsupervised learning. ITL criteria take into consideration the higher-order statistical behavior of the systems and signals as desired. ITL is conceptually different from other kernel methods as it is based on kernel density estimation (KDE) and thus its kernel function need not be positive definite, instead satisfying a different set of properties as detailed in \cite{Parzen1962}. Nevertheless, the estimators in both learning schemes share many similarities \cite{Xu08}, including several positive-definite kernels such as the Gaussian kernel and the Laplacian kernel \cite{Parzen1962}. In fact, positive definiteness is preferred in ITL due to numerical stability in computation.

In the standard kernel method approach, points in the input space $\mathbf{x}_i\in\mathcal{X}$ are mapped, using an implicit nonlinear function $\phi(\cdot)$, into a potentially infinite-dimensional inner product space or RKHS, denoted by $\mathcal{H}$. The explicit representation is of secondary nature. The Mercer condition guarantees the existence of the mapping. A real valued similarity function $k:\mathcal{X}\times\mathcal{X}\rightarrow \R$ is defined as
\begin{equation}
k(\mathbf{x},\mathbf{x}')=\langle\phi(\mathbf{x}),\phi(\mathbf{x}')\rangle
\end{equation}
which is referred to as a reproducing kernel. This presents an elegant solution for classification, clustering, regression, and principal component analysis, since the mapped data points are linearly separable in the potentially infinite-dimensional RKHS, allowing classical linear methods to be applied directly on the data. However, because the actually points (functions) in the function space are inaccessible, kernel methods scale poorly to large datasets. Naive kernel methods operate on the kernel or Gram matrix, whose entries are denoted $\mathbf{K}_{i,j}=k(\mathbf{x}_i,\mathbf{x}_j)$, requiring $O(N^2)$ space complexity and $O(N^3)$ computational complexity for many standard operations. For online kernel adaptive filtering (KAF) algorithms \cite{Liu10,KAARMA,Li2018,Li2019functional}, this represents a rolling sum with linear or superlinear growth. There have been a continual effort to sparsify and reduce the computational load, especially for online KAF \cite{QKLMS, NICE,SNIPGOAL}.

The two most important concepts in ITL are the information potential (IP), which is associated with R\'{e}nyi's quadratic entropy (QE), and the cross information potential (CIP) that measures dissimilarity between two density functions \cite{Xu08}. The estimator of IP requires summing all the elements of the kernel or Gram matrix. A straightforward computation is expensive in both storage and time, especially when the number of samples $N$ is large. Different methods have been proposed to reduce this computational burden by extracting relevant information with sufficient accuracy without processing all $N^2$ elements of the Gram matrix \cite{SmolaSparseGreedy00, WilliamsIDD2000, Fine02, Seth09}.

\begin{figure}[t!]
	\centering
	\includegraphics[width=0.30\textwidth]{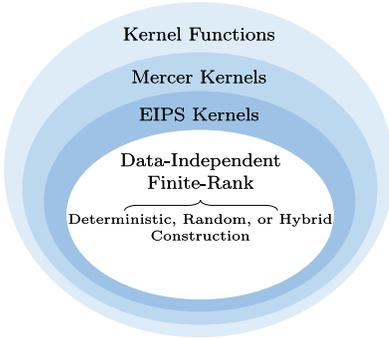}
	\caption{A taxonomy of kernels. Kernel functions for kernel density estimation (KDE) need not be positive definite. Here, we focus on the family of positive-definite explicit inner product space (EIPS) kernels.}
	\label{fig:tax}
\end{figure}

Recently, we proposed a no-trick (NT) framework for kernel adaptive filtering (KAF) using explicit feature mappings that define a positive definite kernel for a finite-dimensional RKHS \cite{Li2019notrick}. The same concept can be integrated seamlessly into ITL using a family of estimators based on separable finite-rank or degenerate kernels whose basis are sampled or constructed independently of the training data. Instead of manipulating the data through pruning or sparsification, we design a family of finite-rank explicit inner produce space (EIPS) Mercer kernels, specifically their explicit feature mappings, for fast, scalable, and accurate estimators for ITL. The Mercer theorem states: 
\begin{thm}[Mercer kernel]
	\label{defn:MercerKernel}
	Let $\mu$ be a probability measure on $\mathcal{X}$, and $\mathcal{H}_{\mu}(\mathcal{X})$ the associated Hilbert space. Given a sequence $(\lambda_i)_{i\in\mathbb{N}}\in\ell^1$ with $\lambda_i\geq 0$, and an orthogonal family of unit norm functions $(\psi_i)_{i\in\mathbb{N}}$ with $\psi_i\in\mathcal{H}_{\mu}(\mathcal{X})$, the associated Mercer kernel is
	\begin{equation}
	k(\x,\x')=\sum^\infty_{i=1}\lambda_i\psi_i(\x)\psi_i(\x')\label{eq:GeneralKernel}
	\end{equation}
	where $\lambda_i$ are the eigenvalues of the kernel and $\psi_i$ its eigenfunctions, and the series' convergence is absolute and uniform.
\end{thm}
In practice, for simplicity, a Mercer kernel where the infinite sum in \eqref{eq:GeneralKernel} can be expressed in closed form is often used, e.g., the Gaussian kernel function, and the expansion itself is either unknown or ignored. In this paper, we take an alternative approach and focus on the family of EIPS kernel functions (specifically data-independent finite-rank) shown in Fig. \ref{fig:tax}, that accelerates the computation of ITL quantities with the utmost versatility and convenience. Compared to \eqref{eq:GeneralKernel} which consists of a continuous orthonormal basis of eigenfunctions, finite-rank or degenerate Mercer kernels of rank $D\geq 1$ are expressed using finite series
\begin{equation}
k(\x,\x') =  \sum_{i=1}^{D}\lambda_i \psi_i(\x)\psi_i(\x').
\end{equation}

Defining an EIPS allows a weighted sum of finite-rank kernel evaluations to be factorized and collapsed for later use as a consolidated feature vector. This is especially efficient when coupled with KAF using ITL cost functions such as the maximum correntropy criterion (MCC) and the minimum error entropy (MEE). Other ITL estimators such as that of Cauchy-Schwartz quadratic mutual information (QMI-CS) and Euclidean distance based quadratic mutual information (QMI-ED) also benefit from the reduced computational complexity offered by this family of fast, scalable, and accurate ITL estimators.

\subsection{Related Work}

A related concept is the fast multipole method (FMM) \cite{GREENGARD1987FMM}, developed for the rapid summation of potential fields generated by a large number of sources (N-body problem in mathematical physics), in which the potential function is expanded in multipole (singular) series and local (regular) series at the expansion centers. This typically combines a far-field expansion of the kernel, in which the influence of sources and targets separates, with a hierarchical subdivision of space of sources into panels or clusters. For the Gaussian field, various factorization and space subdivision schemes include the fast Gauss transform (FGT) and the improved Gauss transform \cite{Greengard1991}. The improved FGT for KDE uses the greedy farthest-point clustering algorithm to model the space subdivsion task as a $\mathtt{k}$-center problem \cite{Yang2003}. Unfortunately, their effectiveness diminishes for higher dimensions and large datasets, since Hermite expansion is used for FGT, resulting in $p^d$ terms for a $p$-term truncation in $d$ dimensions, i.e., exponential growth in the accumulation of expansion products along each data dimension. The improved FGT uses multivariate Taylor series (TS) expansion to reduce the number of expansion terms to polynomial order. 

Here, we take the no-trick (NT) kernel method interpretation in \cite{Li2019notrick} by defining an EIPS Mercer kernel equal to the scalar or inner product of the transformed points in an higher finite-dimensional RKHS $\mathcal{H}'$ using explicit mapping $z(\cdot)$, i.e., $k'(\x-\x')=\langle z(\x),z(\x')\rangle_{\mathcal{H}'}$. Mercer condition guarantees the existence of the underlying mapping and universal approximation. From the inner product perspective, an EIPS kernel naturally factorizes the pairwise interaction between two feature vectors, yielding fast, scalable, and accurate solutions, without the computational overhead of clustering the sources. Compared to FMM, the EIPS approach goes further in the abstraction, by defining an equivalent positive-definite kernel (where the inner product between two points are computed using the explicitly mapped feature vectors), therefore, it is not merely an approximation method, but rather, a new, exact kernel formulation within the unifying framework of the RKHS. In this paradigm, the linear combination (sum) of the training data (source points) feature vectors is a linear function represented by a weight vector in this space. Furthermore, in applications such as KAF, we are always interested in following the embedded trajectory of the input signal (local approximation to the trajectory), so we do not need to seek expansions in other parts of the space, unlike FMM. The EIPS kernel method is both efficient and effective for low-dimensional KDE, e.g., an EIPS-ITL estimator for information quantities based on the prediction error, which is typically one dimensional for time series prediction, extracts more information than $2\textsuperscript{nd}$-order statical models such as \cite{Li13KAFCI}. Without loss of generality, we will use the simple TS expansion EIPS kernel as ITL estimator in low dimensions. For higher dimensions, we will instead use Gaussian quadrature (GQ) with subsampled grids to directly control the number of features used in the feature mapping or EIPS kernel, which has been shown to be effective for high dimensional and large data \cite{Dao2017}.

Random Fourier features (RFF) \cite{rahimi2007RFF} have been successfully applied for efficient kernel learning using finite-rank kernels. While RFF belong to the EIPS family (whose basis are sampled randomly and independent of the training data), for small dimensions, deterministic maps yield significantly lower error and performance variance. For higher dimensions, they can also produce inferior results compared to deterministic polynomial-exact based sampling method, e.g., for online kernel adaptive filtering \cite{Li2019notrick}. Nonetheless, they represent a simple and efficient way to construct EIPS kernels.

Low-rank approximation methods such as the Nystr\"{o}m method \cite{Williams00Nystrom} (basis functions are randomly sampled from the training examples) are data dependent, making them less appealing than data-independent EIPS method. The incomplete Cholesky decomposition (ICD) is another data-dependent approximation method that has been shown to speed up the computation of information theoretic quantities with state-of-the-art ITL performances, with state-of-the-art ITL performances, by leveraging the fact that the eigenvalues of the Gram matrix $\mathbf{K}$ diminishes rapidly and can be replaced by a lower ranked approximation \cite{Bach05ICD, Seth09}. The $N \times N$ symmetric positive definite matrix $K$ can be expressed as $\mathbf{K}=\mathbf{G}\mathbf{G}^\intercal$, where $\mathbf{G}$ is an $N\times N$ lower triangular matrix with positive diagonal entires, a special case of $LU$ decomposition. Using a greedy approach, the ICD minimizes the trace (sum of eigenvalues) of the residual $\norm{\mathbf{K}-\tilde{\mathbf{G}}\tilde{\mathbf{G}}^\intercal}<\epsilon$ with an $N\times D$ (where $D\leq N$) lower triangular matrix $\tilde{\G}$ with arbitrary accuracy, where $\epsilon$ is a small positive number of choice and $\norm{\cdot}$ is a suitable matrix form. The value of $D$, which determines the space and time complexity, $O(ND)$ and $O(ND^2)$, respectively, is indirectly set by the desired precision $\epsilon$, depending on the density of the samples. Furthermore, computing the decomposition using $\tilde{\G}$ is not only a data-dependent batch method, but also comes with considerable computational overhead. They still require computing the kernel matrix. The EIPS-ITL estimators, on the other hand, is a full kernel approach that defines a positive-definite kernel using explicitly mapped features from data-independent basis. The feature space dimension $D$ is set directly, allowing greater control in resource allocation and simplified implementation, especially for online applications.

The rest of the paper is organized as follows. In Section \ref{Sec:RKHS}, explicit-inner-produce-space kernel construction is discussed. Information theoretic learning is reviewed in Section \ref{Sec:NTITL}, and EIPS-ITL estimators are presented. Experimental results are shown in Section \ref{Sec:Simulation}. Finally, Section \ref{Sec:Conclusion} concludes this paper.

\section{EIPS Feature Mapping Construction}\label{Sec:RKHS}
To accelerate ITL estimators, we propose to map the input data to a higher finite-dimensional feature space using EIPS features. Having data-independent basis improves its versatility significantly, allowing the mapping to be predetermined and implemented online with greater efficiency. The explicit feature mapping can be constructed either deterministically, randomly, or via a combination of the two approaches (hybrid). These mappings define a new, equivalent reproducing kernel with universal approximation property \cite{Li2019notrick}. Furthermore, the inner product in the finite-dimensional RKHS naturally factorizes the pairwise interactions and greatly simply the computation and storage of ITL quantities, e.g., they reduce the cost of computing all pairwise interactions for $N$ points from $O(N^2)$ to $O(N)$ and consolidate the collection of $N$ points into a single weight vector of dimension $D\ll N$. 

The popular random Fourier features \cite{rahimi2007RFF} belong to a class of randomly constructed EIPS kernels for scaling up kernel machines. The underlying principle states:
\begin{thm}[Bochner, 1932\cite{Bochner1959}]
	A continuous shift-invariant properly-scaled kernel $k(\mathbf{x},\mathbf{x}')=k(\mathbf{x}-\mathbf{x}'):\R^d\times\R^d\rightarrow\R$, and $k(\x,\x)=1,\forall\x$, is positive definite if and only if $k$ is the Fourier transform of a proper probability distribution.
\end{thm}
The corresponding kernel can then be expressed in terms of its Fourier transform $p(\bm{\omega})$ (a probability distribution) as
%and defining $\zeta_{\bm{\omega}}(\mathbf{x})=e^{j\bm{\omega}^\intercal\mathbf{x}} $, the corresponding shift-invariant kernel can be written as
\begin{align}
k(\mathbf{x}-\mathbf{x}')&=\int_{\R^d}p(\bm{\omega})e^{j\bm{\omega}^\intercal(\mathbf{x}-\mathbf{x}')}d\bm{\omega}\label{eq:kintergral}\\
&={\rm E}_{\bm{\omega}}\left[e^{j\bm{\omega}^\intercal(\mathbf{x}-\mathbf{x}')}\right]={\rm E}_{\bm{\omega}}\left[\langle e^{j\bm{\omega}^\intercal\mathbf{x}},e^{j\bm{\omega}^\intercal\mathbf{x}'}\rangle\right]\label{eq:Fourier_kernel}
\end{align}
where $\langle \cdot,\cdot \rangle$ is the Hermitian inner product $\langle \mathbf{x},\mathbf{x}'\rangle =\sum_i\mathbf{x}_i\overline{\mathbf{x}'_i}$, and $\langle e^{j\bm{\omega}^\intercal\mathbf{x}},e^{j\bm{\omega}^\intercal\mathbf{x}'}\rangle$ is an unbiased estimate of the properly scaled shift-invariant kernel $k(\mathbf{x}-\mathbf{x}')$ when $\bm{\omega}$ is drawn from the probability distribution $p(\bm{\omega})$. We ignore the imaginary part of the complex exponentials to obtain a real-valued mapping.

Alternatively, the RFF approach can be viewed as performing numerical integration using randomly selected sample points. In numerical analysis, there are many polynomial-exact ways to approximate the integral with a discrete sum of judiciously selected points. For small input dimensions, deterministic feature mappings, such as Taylor series expansion, yield significantly lower error and performance variance than random maps. For data of higher dimensions, polynomial-exact deterministic features can be sampled from the distribution determined by their weights to combat the curse of dimensionality and gain direct control over the feature dimension. We have analyzed the performances of deterministic vs. random features for online kernel adaptive filtering in \cite{Li2019notrick}. In this paper, we will briefly summarize the class of deterministically constructed EIPS kernel for ITL estimators.

\subsection{Taylor Polynomial Features}\label{sec:taylor_features}
This is the most straightforward deterministic feature map for EIPS based on the Gaussian kernel, where each term in the TS expansion is expressed as a sum of matching monomials in the data pair $\mathbf{x}$ and $\mathbf{x}'$, i.e.,
\begin{equation}\label{eq:express_Gaussian}
k(\mathbf{x}, \mathbf{x}') = e^{-\frac{\|\mathbf{x}- \mathbf{x}'\|^2}{2\sigma^2}} =
e^{-\frac{\|\mathbf{x}\|^2}{2\sigma^2}}e^{-\frac{\|\mathbf{x}'\|^2}{2\sigma^2}}e^{\frac{\langle \mathbf{x}, \mathbf{x}'\rangle}{\sigma^2}}.
\end{equation}
We can easily factor out the product terms that depend on $\x$ and $\x'$ independently. The joint term in \eqref{eq:express_Gaussian}, $e^{\frac{\langle \mathbf{x}, \mathbf{x}'\rangle}{\sigma^2}}$, can be expressed as a power series or infinite sum using Taylor polynomials as
\begin{equation}\label{eq:scalar_taylor}
e^{\frac{\inner{\x,\x'}}{\sigma^2}} = \sum_{n=0}^{\infty}
\frac{1}{n!}\left(\frac{\inner{\x,\x'}}{\sigma^2}\right)^n.
\end{equation}
Using shorthand, we can factor the inner-product exponentiation as
\begin{equation}\label{eq:multinomial_exp}
\inner{\x,\x'}^n=\left(\sum_{i=1}^d \x_i \x'_i\right)^n=\sum_{j \in [d]^n} \left(
\prod_{i=1}^n \x_{j} \right) \left( \prod_{i=1}^n \x'_{j} \right)
\end{equation}
where $j$ enumerates over all selections of $d$ coordinates (including repetitions and different orderings of the same coordinates) thus avoiding collecting equivalent terms and writing down their corresponding multinomial coefficients, i.e., as an inner product between degree $n$ monomials of the coordinates of $\x$ and $\x'$. Substituting this into \eqref{eq:scalar_taylor} and \eqref{eq:express_Gaussian} yields the following explicit feature map:
\begin{equation}
\label{eq:taylor_features} z_{n,j}\left( \x \right) =
e^{-\frac{\|\x\|^2}{2\sigma^2}} \frac{1}{\sigma^{n}\sqrt{n!}}\prod_{i=0}^{n}
{\x}_{j}
\end{equation}
where $k(\x,\x') = \inner{z(\x),z(\x')} = \prod_{k=0}^{\infty} \prod_{j\in [d]^k} z_{n,j}(\x)z_{n,j}(\x')$. For TS feature approximation or EIPS kernel construction, we truncate the infinite sum to the first $r+1$ terms:
\begin{equation}
\label{eq:taylor_kernel_expansion} k'(\x,\x')=\inner{z_r(\x),z_r(\x')}
= e^{-\frac{\|\x\|^2+\|\x'\|^2}{2\sigma^2}} \sum_{n=0}^{r} \frac{1}{n!}\left(\frac{\inner{\x,\x'}}{\sigma^2}\right)^n
\end{equation}
where the TS approximation is exact up to polynomials of degree $r$.

In practice, the different permutations of j in each $n$-th term of Taylor expansion, \eqref{eq:multinomial_exp}, can be grouped into a single feature corresponding to a distinct monomial, resulting in ${{d+n-1} \choose {n}}$ features of degree $n$, and a total of $D={{d+r} \choose {r}}$ features of degree at most $r$.
\subsubsection{Precision of Taylor Series Expansion}\label{Subsec:TSPrecision}
For the Gaussian kernel, the precision of TS expansion can be defined precisely using the mean-value form of the approximation remainder.
\begin{thm}[Taylor's Formula]
	Let the function $f:\R\rightarrow\R$ be $r+1$ times differentiable (where the integer $r\geq 1$) on the open interval with $f^{(r)}$ continuous on the closed interval between $x_0$ and $x$, the remainder $R_r$ of the Taylor polynomial is
	\begin{align}
	R_r(x)=\frac{f^{(r+1)}(\xi_x)}{(r+1)!}(x-x_0)^{r+1}\label{eq:TSremainder}
	\end{align}
	for some real number $\xi_x$ between $x_0$ and $x$.
\end{thm} 
Suppose we want the desired accuracy to be within $10^{-6}$ in absolute error, i.e., $|e^x-P_r(x)|\leq 10^{-6}$, for all $x\in[-1,1]$, where $P_r(x)=\sum_{n=0}^{r}\frac{f^{(n)}(x_0)}{n!}(x-x_0)^n$ is the $n$-th order Taylor polynomial. Solving for the worst case, $\frac{e}{(r+1)!}<10^{-6}$, we have $r\geq 9$. In Section \ref{Subsec:Accelerate}, we will illustrate this by comparing the performance of Taylor polynomial EIPS formulation with state-of-the-art ITL reduced-rank-approximation fast method using incomplete Cholesky decomposition. 

\subsection{Gaussian Quadrature (GQ) Features with Subsampled Grids}
A \emph{quadrature rule} is a choice of points $\omega_i$ and weights $a_i$ to minimize the maximum error $\epsilon$. For a fixed diameter $M$, the \emph{sample complexity} (SC) is defined as:
\begin{defn}
	\label{defnSampleComplexity}
	For any $\epsilon > 0$, a quadrature rule has sample complexity $D_{\mathrm{SC}}(\epsilon) = D$, where $D$ is the smallest number of samples such that the rule yields a maximum error of at most $\epsilon$.
\end{defn}

There are many quadrature rules, without loss of generality, we focus on Gaussian quadrature (GQ), specifically the Gauss-Hermite quadrature using Hermite polynomials. In numerical analysis, GQ is an exact-polynomial approximation of a one-dimensional definite integral:  $\int p(\bm{\omega}) f(\bm{\omega}) \, d \bm{\omega} \approx \sum_{i=1}^D a_i f(\bm{\omega}_i)$, where the $D$-point construction yields an exact result for polynomials of degree up to $2D - 1$. While the GQ points and corresponding weights are both distribution $p(\bm{\omega})$ and parameter $D$ dependent, they can be computed efficiently using orthogonal polynomials. GQ approximations are accurate for integrating functions that are well-approximated by polynomials, including all sub-Gaussian densities. Compared to random Fourier features, GQ features have a much weaker dependence on the approximation error $\epsilon$, at a constant cost of an additional factor of $2^d$ (independent of the error $\epsilon$) \cite{Dao2017}.

To extend one-dimensional GQ to higher dimensions, grid-based quadrature rules can be constructed efficiently. A \emph{dense grid} or tensor-product construction factors the integral \eqref{eq:kintergral} along the dimensions
$k(u) = \prod_{i=1}^d \left(\int_{-\infty}^{\infty} p_i(\omega)
\exp(j \omega e_i^\intercal u) \, d \omega \right)$, where $e_i$ are the standard basis vectors, and can be approximated using one-dimensional quadrature rule. However, since the sample complexity is doubly-exponential in $d$, a \emph{sparse grid} or Smolyak quadrature is typically used \cite{Smolyak63}. Only points up to some fixed total level A is included, achieving a similar error with exponentially fewer points than a single larger quadrature rule. 

The major drawback of the grid-based construction is the lack of fine tuning for the feature dimension. Since the number of samples extracted in the feature map is determined by the degree of polynomial exactness, even a small incremental change can produce a significant increase in the number of features. \emph{Subsampling} according to the distribution determined by their weights is used to combat both the curse of dimensionality and the lack of detailed control over the exact feature number. There are also data-adaptive methods to choose a quadrature rule for a predefined number of samples \cite{Dao2017}, but we are focused on data-independent EIPS features.

\subsection{Universal Approximation}
EIPS feature mappings such as random Fourier features, Gaussian quadrature, and Taylor polynomials are not only an approximation method, but also defines an equivalent kernel that induces a new reproducing kernel Hilbert space: a nonlinear mapping $z(\cdot)$ that transforms the data from the original input space to a new higher finite-dimensional RKHS $\mathcal{H}'$ where $k'(\x-\x')=\langle z(\x),z(\x')\rangle_{\mathcal{H}'}$. The RKHS $\mathcal{H}'$ is not necessarily contained in the RKHS $\mathcal{H}$ corresponding to the kernel function $k$, e.g., Gaussian kernel. It is easy to show that the EIPS mappings discussed in this paper induce a positive-definite kernel function satisfying Mercer's conditions.

\begin{pro}[Closure properties]
	Let $k_1$ and $k_2$ be positive-definite kernels over $\mathcal{X}\times\mathcal{X}$ (where $\mathcal{X}\subseteq\R^d$), $a\in \R^+$ is a positive real number, $f(\cdot)$ a real-valued function on $\mathcal{X}$, then the following functions are positive definite kernels.
	\begin{enumerate}
		\item $k(\x,\x') = k_1(\x,\x')+k_2(\x,\x')$,
		\item $k(\x,\x') = ak_1(\x,\x')$,
		\item $k(\x,\x') = k_1(\x,\x')k_2(\x,\x')$,
		\item $k(\x,\x') = f(\x)f(\x')$.
	\end{enumerate}
\end{pro}
Since exponentials and polynomials are positive-definite kernels, under the closure properties, it is clear that the inner products of random Fourier features, Gaussian quadrature, and Taylor polynomials are all reproducing kernels. It follows that these kernels have universal approximating property: approximates uniformly an arbitrary continuous target function to any degree of accuracy over any compact subset of the input space.

\section{EIPS kernel for Information Theoretic Learning (EIPS-ITL)}\label{Sec:NTITL}
ITL is a framework to adapt nonparametric systems using information quantities such as entropy and divergence \cite{ITL}. ITL criteria is still directly estimated from data via Parzen kernel estimator, but it extracts more information from the data for adaptation, and yields, therefore, solutions that are more accurate than mean squared error (MSE) in non-Gaussian and nonlinear signal processing. Reproducing kernels are covariance functions explains their early role in inference problems \cite{Aronszajn1950,Parzen1959}. R\'{e}nyi's quadratic entropy of a random variable $X$ with pdf $f_X(x)$ is defined as
\begin{align}
H_2(X) \stackrel{\Delta}{=} -\log\int f^2_X(x)dx.
\end{align}
The Parzen estimate of the pdf, given a set of independent and identically distributed (i.i.d.) data $\{x_i\}^N_{i=1}$ drawn from the distribution is 
\begin{align}
\hat{f}_{X;\sigma}(x)=\frac{1}{N}\sum^{N}_{i=1}\mathcal{K}_\sigma(x-x_i)\label{eq:ParzenPDF}
\end{align}
where $N$ is the number of data samples, and $\mathcal{K}_\sigma$ is the Gaussian kernel with kernel size $\sigma$
\begin{align}
\mathcal{K}_\sigma(x-x_i)=\frac{1}{\sqrt{2\pi}\sigma}\exp\left(-\frac{(x-x_i)^2}{2\sigma^2}\right).
\end{align}
Without loss of generality, we will only consider the Gaussian kernel and related EIPS kernels in this paper.

Using the no-trick or EIPS explicit mapping $z(\cdot)$, the kernel function in \eqref{eq:ParzenPDF} is replaced with the inner product of the explicitly mapped points (functions) in the finite-dimensional RKHS $\mathcal{H}'$ as
 \begin{align}
 \tilde{f}_{X;\sigma}(z(x))&=\frac{1}{N}\sum^{N}_{i=1}z(x)^\intercal\cdot z(x_i)\nonumber\\
 &=\frac{z(x)^\intercal}{N}\sum^{N}_{i=1} z(x_i)=z(x)^\intercal\overbar{z(x)}
 \label{eq:NTParzenPDF}
 \end{align}
 where $\overbar{z(x)}=(1/N)\sum^{N}_{i=1}\cdot z(x_i)$ is the sample mean or centroid and is, in general, independent of the target $x$ or $z(x)$. Alternatively, from the RKHS paradigm, this can be viewed as a weight vector that represents or parametrizes the linear function in the EIPS, i.e.,  $\tilde{f}_{X;\sigma}(\cdot)=\langle \overbar{z(x)},\cdot\rangle$. 

A nonparametric estimate of R\'{e}nyi's quadratic entropy directly from samples is
\begin{align}
\hat{H}_2(X)=-\log {\rm IP}(X)
\end{align}
where the information potential (IP) is defined as
\begin{align}
{\rm IP}(X)\stackrel{\Delta}{=}\frac{1}{N^2}\sum^{N}_{i=1}\sum^{N}_{j=1}\mathcal{K}_{\sqrt{2}\sigma}(x_i-x_j).
\end{align}

Using EIPS \eqref{eq:NTParzenPDF}, the IP estimate becomes
\begin{align}
{\rm IP}(X)&\stackrel{\Delta}{=}\frac{1}{N^2}\sum^{N}_{i=1}\sum^{N}_{j=1}\mathcal{K}_{\sqrt{2}\sigma}(x_i-x_j)\nonumber\\
&=\frac{1}{N^2}\sum^{N}_{i=1}\sum^{N}_{j=1}z(x_i)^\intercal z(x_j)\nonumber\\
&=\underbrace{\frac{1}{N}\sum^{N}_{i=1}z(x_i)^\intercal}_{\overbar{z(x)}^\intercal}\underbrace{\frac{1}{N}\sum^{N}_{j=1}z(x_j)}_{\overbar{z(x)}}\nonumber\\
&=\overbar{z(x)}^\intercal\overbar{z(x)}.
\end{align}
This drastically reduces the quadratic complexity from $O(N^2)$ to a linear $O(N)$, which only requires computing the weight vector or center $\overbar{z(x)}$ once, then squaring it, i.e., scalar product with $O(1)$. Online update of this term is embarrassingly simple, as new sources or sample points are simply added to the existing weight vector with the appropriate normalization, i.e., $\overbar{z(x)}_{N+1} = (N\overbar{z(x)}_{N} + z(x_{N+1}))/(N+1)$.

Let $\{X_t,t\in T\}$ be a stochastic process with $T$ being an index set. The nonlinear mapping $\phi$ induced by the Gaussian kernel maps the data into the feature space $\mathbb{F}$, where the auto-correntropy function $V_X(t,t+\tau)$ is defined from $T\times T$ into $\mathbb{R}^+$ given by
\begin{align}
V_X(t,t+\tau)&\stackrel{\Delta}{=}\Ex[\langle\phi(X_t),\phi(X_{t+\tau})\rangle_\mathbb{F}]\\
&=\Ex[\mathcal{K}_{\sigma}(X_t-X_{t+\tau})]
%V_\sigma(X,Y) \stackrel{\Delta}{=} \mathrm{E}[\mathcal{K}_{\sigma}(X-Y)]
\end{align}
where $\Ex[\cdot]$ denotes the expectation. A sufficient condition for $V(t,t-\tau) = V(\tau)$ is that the stochastic process must be strictly stationary on all the even moments, a stronger condition than wide sense stationarity (limited to \nth{2} order moments). The IP is the mean squared projected data $\langle \frac{1}{N}\sum^N_{i=1}\phi(x_i),\frac{1}{N}\sum^N_{j=1}\phi(x_j) \rangle$or the expected value of correntropy over lags $\tau$. A more general form of correntropy (cross-correntropy) \cite{LiuCorrentropy2007} between two random variables is defined as
\begin{align}
V_\sigma(X,Y) \stackrel{\Delta}{=} \Ex[\mathcal{K}_{\sigma}(X-Y)].
\end{align}
The sample estimate of correntropy for a finite number of data $\{(x_i,y_i)\}^N_{i=1}$ is
\begin{align}
\hat{V}_{N,\sigma}(X,Y)=\frac{1}{N}\sum^N_{i=1}\mathcal{K}_{\sigma}(x_i-y_i).
\end{align}
Using Taylor series expansion for the Gaussian kernel, correntropy can be expressed as
\begin{align}
V_{\sigma}(X,Y)=\frac{1}{\sqrt{2\pi}\sigma}\sum^\infty_{n=0}\frac{(-1)^n}{2^n\sigma^{2n}n!}\Ex[(X-Y)^{2n}]
\end{align}
which involves all the even-order moments of the random variable $X-Y$ (where the kernel choice dictates the expansion, e.g., the sigmoidal kernel contains all the odd moments) \cite{Santamaria2006}. 

In fact, all learning algorithms that use nonparametric pdf estimates in the input space admit an alternative formulation as kernel methods expressed in terms of inner products. As shown above, the kernel techniques are able to extract higher order statistics of the data that should lead to performance improvements for non-Gaussian environments. Next, we show the explicit EIPS derivations of several commonly used ITL estimators.
\subsubsection{EIPS Quadratic Mutual Information (QMI)}
The Cauchy-Schwartz quadratic mutual information and Euclidean distance based QMI are defined, respectively, as
\begin{equation}
I_{CS}\stackrel{\Delta}{=}\log \frac{\int\int f^2_{XY}(x,y)dxdy\int\int f^2_{X}(x)f^2_Y(y)dxdy}{\left(\int\int f_{XY}(x,y)f_X(x)f_Y(y)dxdy\right)^2}
\end{equation}
\begin{align}
I_{ED} &\stackrel{\Delta}{=}\int\int f^2_{XY}(x,y)dxdy+\int\int f^2_{X}(x)f^2_Y(y)dxdy\nonumber\\
& \quad -2\int\int f_{XY}(x,y)f_X(x)f_Y(y)dxdy.
\end{align}
The above expressions consists of the following three distinct terms. The EIPS IP estimate of the joint pdf ($V_J$) is computed as
\begin{align}
\lefteqn{V_J=\int\int \hat{f}^2_{XY}(x,y)dxdy}\nonumber\\
& & & & &= \frac{1}{N^2}\sum_{i=1}^{N}\sum_{j=1}^{N} k(x_i-x_j)k(y_i-y_j)\nonumber\\
& & & & &\stackrel{(a)}{=}\frac{1}{N^2}\sum_{i=1}^{N}\sum_{j=1}^{N} z(x_i)^\intercal z(x_j)z(y_j)^\intercal z(y_i)\nonumber\\
& & & & &\stackrel{(b)}{=}\frac{1}{N^2}\sum_{i=1}^{N}z(x_i)^\intercal\underbrace{\left(\sum_{j=1}^{N}z(x_j)z(y_j)^\intercal\right)}_{\mathbf{Z}_{XY}\stackrel{\Delta}{=}z(X)z(Y)^\intercal} z(y_i)\nonumber\\
& & & & &=\frac{1}{N^2}\sum_{i=1}^{N}z(x_i)^\intercal  \mathbf{Z}_{XY} z(y_i)\nonumber\\
& & & & &=\frac{1}{N^2}\sum_{k=1}^{D}\sum_{l=1}^{D}\left(\left(\sum_{i=1}^{N}z(x_i) z(y_i)^\intercal\right)\odot \mathbf{Z}_{XY}\right)_{k\ell}\nonumber\\
& & & & &=\frac{1}{N^2}\sum_{k=1}^{D}\sum_{\ell=1}^{D}\Big(\mathbf{Z}_{XY}\odot \mathbf{Z}_{XY}\Big)_{k\ell}.
\end{align}
where $(a)$ uses the shift-invariant property, $(b)$ is due to the associative property, where $z(X)=[z(x_1),\cdots z(x_n)]$ with the square matrix $\mathbf{Z}_{XY}\in\R^{D\times D}$, and $\odot$ is the Hadamard product operator. The EIPS IP estimate of the factorized marginal pdf ($V_M$) becomes
\begin{align}
\lefteqn{V_M = \int\int f^2_{X}(x)f^2_Y(y)dxdy}\nonumber\\
& & & & &= \frac{1}{N^4}\sum_{i=1}^{N}\sum_{j=1}^{N} k(x_i-x_j)\sum_{i=1}^{N}\sum_{j=1}^{N}k(y_i-y_j)\nonumber\\
& & & & &= \frac{1}{N^4}\sum_{i=1}^{N}\sum_{j=1}^{N} z(x_i)^\intercal z(x_j)\sum_{i=1}^{N}\sum_{j=1}^{N}z(y_i)^\intercal z(y_j)\nonumber\\
& & & & &= \underbrace{\frac{1}{N}\sum_{i=1}^{N}z(x_i)^\intercal}_{\overbar{z(x)}^\intercal}\underbrace{\frac{1}{N}\sum_{j=1}^{N}z(x_j)}_{\overbar{z(x)}}\underbrace{\frac{1}{N}\sum_{i=1}^{N}z(y_i)^\intercal}_{\overbar{z(y)}^\intercal}\underbrace{\frac{1}{N}\sum_{j=1}^{N} z(y_j)}_{\overbar{z(y)}}\nonumber\\
& & & & &= \overbar{z(x)}^\intercal \overbar{z(x)}\overbar{z(y)}^\intercal\overbar{z(y)}.
\end{align}
And, the EIPS generalized-cross IP estimate ($V_C$) is
\begin{align}
\lefteqn{V_C = \int\int \hat{f}_{XY}(x,y)\hat{f}_X(x)\hat{f}_Y(y)dxdy}\nonumber\\
& & & & &= \frac{1}{N^3}\sum_{i=1}^{N}\sum_{j=1}^{N} \sum_{k=1}^{N}k(x_i-x_j)k(y_i-y_k)\nonumber\\
& & & & &= \frac{1}{N^3}\sum_{i=1}^{N}\sum_{j=1}^{N} \sum_{k=1}^{N}z(x_i)^\intercal z(x_j)z(y_i)^\intercal z(y_k)\nonumber\\
& & & & &= \frac{1}{N^3}\sum_{i=1}^{N}\sum_{j=1}^{N} z(x_i)^\intercal z(x_j)z(y_i)^\intercal\sum_{k=1}^{N}z(y_k)\nonumber\\
& & & & &\stackrel{(c)}{=} \underbrace{\frac{1}{N}\sum_{j=1}^{N}z(x_j)^\intercal}_{\overbar{z(x)}^\intercal}\underbrace{\frac{1}{N}\sum_{i=1}^{N}z(x_i)z(y_i)^\intercal}_{\overbar{\mathbf{Z}}_{XY}}\underbrace{\frac{1}{N}\sum_{k=1}^{N}z(y_k)}_{\overbar{z(y)}}\nonumber\\
& & & & &=\overbar{z(x)}^\intercal \overbar{\mathbf{Z}}_{XY}\overbar{z(y)}
\end{align}
where $(c)$ is due to the commutative property of summation and the fact that the transpose of a scalar is itself, i.e., $z(x_i)^\intercal z(x_j) = z(x_j)^\intercal z(x_i)$.
\subsubsection{EIPS Divergence and Distance Measures}
The CS divergence and ED divergence are defined as
\begin{align}
D_{CS}&\stackrel{\Delta}{=}\log\frac{\int f^2_X(x)dx\int f^2_Y(y)dy}{\big(\int f_X(x) f_Y(x)dx\big)^2}\\
D_{ED}&\stackrel{\Delta}{=}\int f^2_X(x)dx + \int f^2_Y(y)dy-2\int f_X(x) f_Y(x)dx
\end{align}
respectively, where the cross information potential (CIP) estimated can be computed as
\begin{align}
\int \hat{f}_X(z)\hat{f}_Y(z)dz&=\frac{1}{N^2}\sum_{i=1}^{N}\sum_{j=1}^{N}k(x_i-y_j)\nonumber\\
&=\frac{1}{N^2}\sum_{i=1}^{N}\sum_{j=1}^{N}z(x_i)^\intercal z(y_j)\nonumber\\
&=\underbrace{\frac{1}{N}\sum_{i=1}^{N}z(x_i)^\intercal}_{\overbar{z(x)}^\intercal}\underbrace{\frac{1}{N}\sum_{j=1}^{N}z(y_j)}_{\overbar{z(y)}}\nonumber\\
&=\overbar{z(x)}^\intercal \overbar{z(y)}.
\end{align}
It follows that the correntropy coefficient estimate is
\begin{align}
c&\stackrel{\Delta}{=}\frac{\Ex_{XY}k(X-Y)-\Ex_{X}\Ex_{Y}k(X-Y)}{\sqrt{\big(1-\Ex_{X_1}\Ex_{X_2}k(X_1-X_2)\big)\big(1-\Ex_{Y_1}\Ex_{Y_2}k(Y_1-Y_2)\big)}}\nonumber\\
\hat{c}&=\frac{\frac{1}{N}\sum_{i=1}^N z(x_i)^\intercal z(y_i)-\overbar{z(x)}^\intercal \overbar{z(y)}}{\sqrt{\big(1-\overbar{z(x)}^\intercal \overbar{z(x)}\big)\big(1-\overbar{z(y)}^\intercal \overbar{z(y)}\big)}}.
\end{align}

\subsection{NT Kernel Adaptive Filtering using EIPS-ITL Criteria}\label{Sec:KAF}
EIPS not only facilitates the computation of ITL quantities, but also integrates seamlessly into online kernel adaptive information filters, as it did for no-trick KAF using conventional MSE criterion \cite{Li2019notrick}.
\subsubsection{NT Maximum Correntropy Criterion}\label{Sec:KAF}
The counterpart to the kernel least mean square (KLMS) \cite{KLMS} algorithm, which adopts the MSE as the cost, is the kernel maximum correntropy criterion (KMCC) filter \cite{KMCC}. Second-order statistics may not be suitable for all nonlinear, especially non-Gaussian, situations. The KMCC combines the simplicity of the KLMS with the higher-order statistics of the correntropy criterion. Using the NT formulation, the NT-KMCC is summarized in Alg. \ref{alg:NT-KMCC}. Compared to the NT-KLMS \cite{Li2019notrick}, we can see that the NT-KMCC has a variable step size controlled by the prediction error.

\begin{algorithm}
	%\SetAlgoLined
	%\small
	\textbf{Initialization:}\\
	$z(\cdot):\X\rightarrow\R^D$: NT feature map\\
	$\mathbf{w}(0) = \textbf{0}$: feature space weight vector $\mathbf{w}\in\R^D$\\
	$\eta$: learning rate\\
	\textbf{Computation:}\\
	\For{n = 1, 2, $\cdots$}{
		$e_n = y_{n} -\textbf{w}^\intercal_{n-1}z(\x_{n})$\\
		$\mathbf{w}_{n} = \mathbf{w}_{n-1}+\,\eta \exp(\frac{-e_n^2}{2\sigma^2})e_n z(\x_{n})$
	}
	\normalsize
	\caption{NT-KMCC Algorithm}
	\label{alg:NT-KMCC}	
\end{algorithm}

\subsubsection{EIPS Minimum Error Entropy}\label{Sec:KAF}
Given a batch of $N$ error samples, the information potential estimator using R\'{e}nyi's quadratic entropy is
\begin{equation}
\hat{V}_{2}(e)=\frac{1}{N^2}\sum^N_{i=1}\sum^N_{j=1}k_{\sigma}(e_i-e_j).
\end{equation}
The cost function $J(e)$ for the MEE criterion is given as
\begin{equation}
{\rm MEE}: J(e)=\min_{\mathbf{w}}\hat{V}_{2}(e).
\end{equation}
The IP is smooth and differentiable, to maximum its value, one can simply move in the direction of its gradient
\begin{equation}
\nabla\hat{V}_{2}(e_n)=\frac{1}{N^2}\sum^N_{i=1}\sum^N_{j=1}k'_{\sigma}(e_{n-i}-e_{n-j})(\x_{n-i}-\x_{n-j}).\label{eq:IPGradDoubleSum}
\end{equation}
For online methods, especially KAF where the kernel trick introduces (super)linear complexity, the Gaussian quadratic stochastic information gradient (SIG) is typically used
\begin{equation}
\frac{\partial\hat{V}_{2}(e_n)}{\partial \mathbf{w}_k}=\frac{1}{\sigma^2 L}\sum^{n-1}_{i=n-L}G_{\sigma}(e_n-e_i)(e_n-e_i)(x_n-x_i).
\end{equation}
Using the EIPS approach, the full (expected value or double sum) IP can be computed extremely efficiently. Using shorthand, the explicit feature mapping factorizes the double summation in the full IP gradient \eqref{eq:IPGradDoubleSum} into the following independent terms
\begin{align}
\lefteqn{\sum^N_{i=1}\sum^N_{j=1}z(e_i)z(e_j)(e_i-e_j)(x_i-x_j)=}\nonumber\\
& & &\sum^N_{i=1}z(e_i)e_i x_i\sum^N_{j=1}z(e_j)+\sum^N_{i=1}z(e_i)\sum^N_{j=1}z(e_j)e_j x_j\nonumber\\
& & &-\sum^N_{i=1}z(e_i)e_i \sum^N_{j=1}z(e_j)x_j-\sum^N_{i=1}z(e_i)x_i\sum^N_{j=1}z(e_j)e_j\nonumber\\
& & =&2 \underbrace{\sum^N_{i=1}z(e_i)e_i x_i}_{Z_1}\underbrace{\sum^N_{j=1}z(e_j)}_{Z_2}-2\underbrace{\sum^N_{i=1}z(e_i)e_i}_{Z_3}\underbrace{\sum^N_{j=1}z(e_j)x_j}_{Z_4}\nonumber\\
& &=&2Z_1 Z_2-2Z_3 Z_4 \qquad\qquad\qquad\qquad\qquad\qquad\qquad\label{eq:DVNT}
\end{align}
where the four $Z_i$ scalar terms can be summed independently.
Since the errors $e_i$ are typically small and one dimensional, without loss of generality, we elect to use the simple Taylor series expansion EIPS mapping for $z(e_i)$.

The NT-KMEE is summarized in Alg. \ref{alg:NT-KMEE}. The NT-KMEE-SIG formulation (single sum) follows trivially and can be used to further accelerate online adaptation. Similarly, the self adjusting step-size formulation \cite{HAN20072733} can be easily applied, which scales the step size by a nonnegative factor of $V(0)-V(e_n)$.

\begin{algorithm}
	%\SetAlgoLined
	%\small
	\textbf{Initialization:}\\
	$z(\cdot):\X\rightarrow\R^D$: input NT feature map\\
	$z_e(\cdot):\mathcal{E}\rightarrow\R^{D_e}$: error EIPS feature map\\
	$\mathbf{w}(0) = \textbf{0}$: feature space weight vector $\mathbf{w}\in\R^D$\\
	$\eta$: learning rate\\
	\textbf{Computation:}\\
	\For{n = 1, 2, $\cdots$}{
		$e_n = y_{n} -\textbf{w}^\intercal_{n-1}z(\x_{n})$\\
		$\mathbf{w}_{n} = \mathbf{w}_{n-1}+\,\eta \nabla \hat{V}_2(e_{n})$ \eqref{eq:DVNT}
	}
	\normalsize
	\caption{NT-KMEE Algorithm}
	\label{alg:NT-KMEE}	
\end{algorithm}
\section{Simulation Results}\label{Sec:Simulation}
Extensive comparisons between MSE and MEE techniques have already been performed in \cite{Erdogmus2002,KMCC,KMEE}, here, we will focused on the speed of EIPS kernel framework for ITL.
\subsection{Accelerating ITL Quantities Computation}\label{Subsec:Accelerate}
First, we evaluate the validity of the proposed method using five benchmark datasets from the UCI machine learning repository \cite{UCI_data}. We normalized them individually (iris, cancer, wine, yeast, and abalone) before computing the estimators: z-score followed by scaling the global extrema to $\pm 1$. As all ITL quantities share similar forms, without loss of generality, we computed the Cauchy-Schwartz quadratic mutual information and correntropy coefficient estimates on all possible pairs of features for each dataset (i.e., $x,x'\in\R$), using the direct method, incomplete Cholesky decomposition, and the simple Taylor polynomial EIPS kernel method. The Gaussian kernel size is set at $\sigma = 1/\sqrt{2}$, and the desired precision for ICD is $\epsilon = 10^{-6}$, which corresponds to a minimum of $9$ terms in the TS expansion using \eqref{eq:TSremainder}. Tables \ref{tab:CC} and \ref{tab:QMI-CS} summarize the results averaged over 10 independent trials. The experiments were performed using Intel Core i7-7700 (at 3.60 GHz with 16 GB of RAM) and MATLAB. In each trial, the ITL descriptors' values and CPU times are accumulated over all feature pairs. Since ICD is data-dependent, the average reduced rank $D$ is listed in a separate column. For comparisons, we showed the performances of EIPS kernels using Taylor polynomials of $4$-th (accurate to $10^{-1.6}$) and $9$-th order (accurate to $10^{-6}$), corresponding to $D=5$ and $D = 10$, respectively. As demonstrated in \cite{Seth09}, ICD is able to match the same value as the direct evaluation using $N\times N$ Gram matrices with at least 6-digit accuracy (there is a tiny rounding error for the cancer dataset in the least significant digit after the decimal point, compared to the direct method, as the correntropy coefficients are accumulated over all possible feature pairs in each trial), in a significantly lower computation time. Remarkably, the EIPS method further outperforms the ICD's speed by another order of magnitude (with no accumulated rounding error for the cancer dataset when using $9$-th order TS expansion).

As discussed above, the ICD does not control the space and time complexities directly, i.e., the reduced dimension $D$ cannot be fixed \textit{a priori}. The ICD is useful only when the eigenvalues of the matrix drop sufficiently fast and the original Gram matrix can be represented by a low rank approximation with sufficient accuracy. However, if this ideal condition fails to exist, e.g., if the dimensionality increases with respect to the number of samples, the ICD performance will suffer. The EIPS approach, on the other hands, defines an equivalent kernel function, as such, it is not merely an approximation method, but rather, a new, exact kernel formulation within the theoretically-grounded unifying framework of the RKHS. 

Not only can we compute ITL quantities with ease and accuracy, but we can also integrate it seamlessly into online KAF algorithms using ITL cost functions, demonstrated next.

\begin{table}[ht]\renewcommand{\arraystretch}{1.5}
	\tiny
	\centering\caption{Average Performance of the Direct and Fast Methods: (Correntropy Coefficient).}
	\label{tab:CC}
	\setlength\tabcolsep{1pt}
	\begin{tabular}{ |r|r r|r c r|r r|r r|}
		\cline{2-10}		
		\multicolumn{1}{c|}{}& \multicolumn{2}{c|}{\scriptsize Direct Method} & \multicolumn{3}{c|}{\scriptsize ICD} & \multicolumn{2}{c|}{{\scriptsize EIPS} ($D = 5$)}& \multicolumn{2}{c|}{{\scriptsize EIPS} ($D = 10$)}\\\hline
		\multicolumn{1}{|c|}{\scriptsize Data} &\scriptsize value  & \scriptsize time &\scriptsize  value  &\scriptsize time & $D$ & \scriptsize value  &\scriptsize time &\scriptsize  value  &\scriptsize time\\
		\multicolumn{1}{|c|}{($n$, dim.)}&\scriptsize & \scriptsize (s) &\scriptsize  &\scriptsize (s) &  & \scriptsize &\scriptsize (s) &\scriptsize  &\scriptsize (s)\\ \hline\hline
		{\scriptsize	iris} (150, 4) &1.747235 &   0.0719 &1.747235  &  0.0079 & \cmmnt{D} 8.3 & 1.746707 &   0.0009& 1.747235 &   0.0009 \\ \hline
		{\scriptsize	wine} (178, 13)&6.466733  &  1.2174 &6.466733 &   0.0464 & \cmmnt{D} 7.9 &6.465304   & 0.0027 &6.466733  &  0.0029\\ \hline
		{\scriptsize	cancer} (198, 32) &112.470020  &  9.9328& 112.470021  &  0.2189 & \cmmnt{D} 6.4 & 112.463802  &  0.0124 & 112.470020  &  0.0133\\ \hline
		{\scriptsize	yeast} (1484, 8) &0.296951 & 30.3389 &0.296951  &  0.0661& \cmmnt{D} 7.4 &0.297262   & 0.0033 &0.296951  &  0.0043\\ \hline
		{\scriptsize	abalone} (4177, 8)  &22.637017 & 328.6687 &22.637017 &   0.0971 & \cmmnt{D} 5.3 &  22.637014  &  0.0058 &   22.637017  &  0.0076\\ \hline
	\end{tabular}
	\normalsize
\end{table}
\begin{table}[ht]\renewcommand{\arraystretch}{1.5}
	\tiny
	\centering\caption{Average Performance of the Direct and Fast Methods: (Cauchy-Schwartz Quadratic Mutual Information).}
	\label{tab:QMI-CS}
	\setlength\tabcolsep{1.4pt}
	\begin{tabular}{ |r|r r|r c r|r r|r r|}
		\cline{2-10}		
		\multicolumn{1}{c|}{}& \multicolumn{2}{c|}{\scriptsize Direct Method} & \multicolumn{3}{c|}{\scriptsize ICD} & \multicolumn{2}{c|}{{\scriptsize EIPS} $(D=5)$}& \multicolumn{2}{c|}{{\scriptsize EIPS} $(D=10)$}\\\hline
		\multicolumn{1}{|c|}{\scriptsize Data} &\scriptsize value  & \scriptsize time &\scriptsize  value  &\scriptsize time & $D$ & \scriptsize value  &\scriptsize time &\scriptsize  value  &\scriptsize time\\
		\multicolumn{1}{|c|}{($n$, dim.)}&\scriptsize & \scriptsize (s) &\scriptsize  &\scriptsize (s) &  & \scriptsize &\scriptsize (s) &\scriptsize  &\scriptsize (s)\\ \hline\hline
		{\scriptsize	iris} (150, 4) & 0.086585  &  0.0615 & 0.086585  &  0.0081 & \cmmnt{D} 7.8 & 0.086538  &  0.0006
		 & 0.086585  &  0.0006\\ \hline		
		{\scriptsize	wine} (178, 13)&0.094259  &  0.9411&0.094259  &  0.0496& \cmmnt{D} 7.3 &0.094239  &  0.0024 &0.094259   & 0.0028\\ \hline
		{\scriptsize	cancer} (198, 32) &0.059147  &  7.0841&0.059147  &  0.2353 & \cmmnt{D} 6.0 &0.059141  &  0.0106&0.059147   & 0.0140\\ \hline		
		{\scriptsize	yeast} (1484, 8) &0.000155 &  23.0459&0.000155  &  0.0709 & \cmmnt{D} 5.5 &0.000155  &  0.0044 & 0.000155    & 0.0049\\ \hline
		{\scriptsize	abalone} (4177, 8)  &0.000237 & 217.0791 &0.000237  &  0.1035 & \cmmnt{D} 5.1 &0.000237  &  0.0052&0.000237  &  0.0082
		\\ \hline
	\end{tabular}
	\normalsize
\end{table}

\subsection{NT Kernel Adaptive Information Filtering with Error Entropy and Error Correntropy Criteria}
Here we perform one-step ahead prediction on the Mackey-Glass (MG) chaotic time series \cite{Mackey77}, defined by the following time-delay ordinary differential equation
\begin{equation*}
\frac{d x(t)}{d t} =\frac{\beta x(t-\tau)}{1+x(t-\tau)^{n}} -\gamma x(t)
\end{equation*}
where $\beta=0.2$, $\gamma=0.1$, $\tau=30$, $n=10$, discretized at a sampling period of 6 seconds using the forth-order Runge-Kutta method, with initial condition $x(t) = 0.9$. Chaotic dynamics are extremely sensitive to initial conditions: small differences in initial conditions yields widely diverging outcomes, rendering long-term prediction intractable, in general. 

The data are standardized by subtracting its mean and dividing by its standard deviation, then scaled by the resulting maximum absolute value to guarantee the sample values are within the range of $[-1,1]$.  A time-embedding or input dimension of $d=7$ is used. The results are averaged over 200 independent trials. In each trial, 2000 consecutive samples with random starting point in the time series are used for training, and testing consists of 200 consecutive samples located in the future. 

\begin{figure}[t!]
	\centering
	\includegraphics[width=0.40\textwidth]{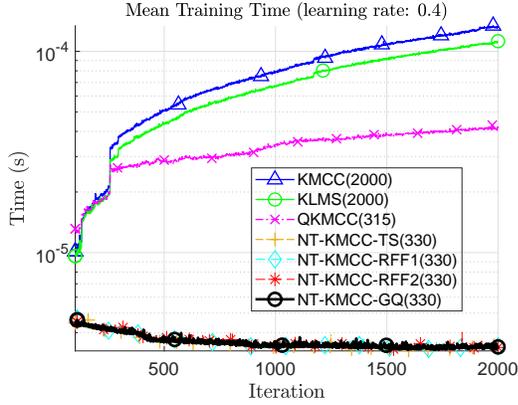}
	\caption{NT-Kernel Maximum Correntropy Criterion (NT-KMCC) algorithm vs. KMCC, and quantized KMCC.}
	\label{fig:KMCC}
\end{figure}

In the first example, we compared the performances of KMCC variants, as shown in Fig. \ref{fig:KMCC}. We fixed the finite-dimensional RKHS dimension for the input features to $D =330$ using $8$-th degree GQ rule with subsampled grids, RFFs (variants 1 and 2 in \cite{Li2019notrick}), and TS expansion. For a comparable resource allocation, we also compared the CPU time with that of the popular vector-quantization sparsification method (QKMCC) with vector quantization parameter set at $q_{\rm factor} = 0.07$ (where the final dictionary size is 315). The Gaussian kernel size is set at $\sigma = 1/\sqrt{2}$, and the learning rate is fixed at $\eta = 0.4$. As expected, comparing to the KLMS, the KMCC requires additional overhead to compute correntropy. The information theoretic computation using EIPS kernels (GQ, RFF1, RFF2, and TS), on the other hand, significantly outperformed the conventional KAF formulations (KLMS and KMCC) and KAF with sparsification (QKMCC) in terms of speed. Again, as is the case for the NT MSE formulations \cite{Li2019notrick}, the average EIPS CPU time is constant across all iterations vs. the (super)linear growth of conventional kernel methods, making EIPS kernel methods ideal for large datasets and continuous online update, e.g., streaming data.

\begin{figure}[t!]
	\centering
	\begin{subfigure}
		\centering
		\includegraphics[width=0.40\textwidth]{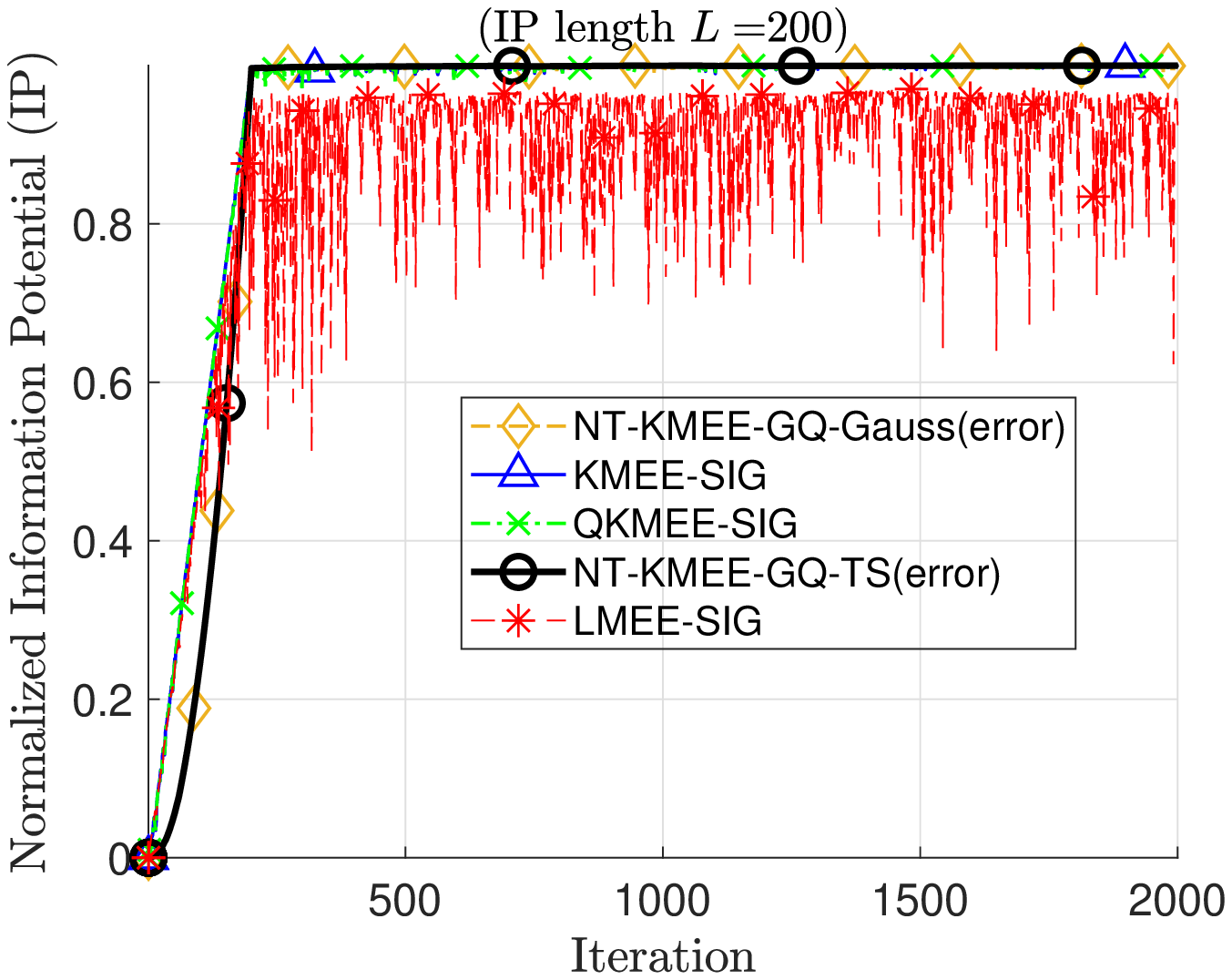}
	\end{subfigure}%
	\begin{subfigure}
		\centering
		\includegraphics[width=0.40\textwidth]{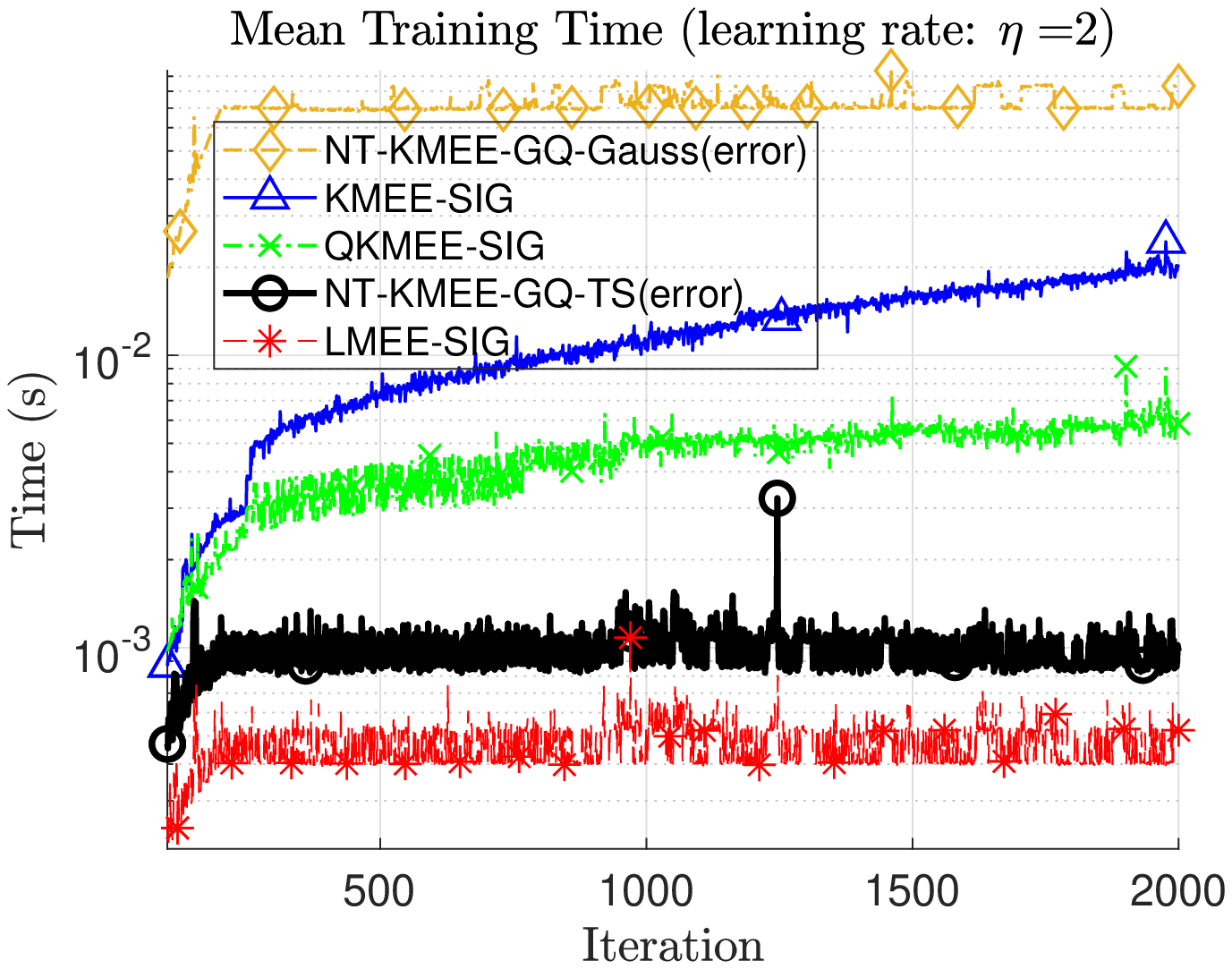}
	\end{subfigure}
	\caption{NT Kernel Minimum Error Entropy (KMEE) algorithms (direct vs. $4$-th order TS expansion for IP) using full information potential (double sum) vs. linear MEE, KMEE, and quantized KMEE with single-sum stochastic information gradient (SIG).}
	\label{fig:KMEE}
\end{figure}

Next, we evaluated the speed of various KMEE implementations in Fig. \ref{fig:KMEE}. IP gradient was computed over the most recent $L=200$ samples or error history. Since extensive comparisons have already been performed between random features and deterministic features for NT KAF in \cite{Li2019notrick}, for clarity of presentation, we focused on the GQ NT formulations. To further showcase the computational efficiency of the EIPS kernel method, we pit the NT-KMEE algorithms using full information potential (double sum) against the linear MEE, kernel MEE, and quantized kernel MEE using the much simpler, linear complexity single-sum stochastic information gradient or SIG. As expected, the direct method to compute the full, expected value of IP, NT-KMEE-GQ-Gauss(error), yielded the worst performance. Nonetheless, we see that the NT kernel method maintains constant complexity. In contrast, the CPU time for conventional kernel method such as KMEE-SIG will continue to increase as the number of training samples (update iterations) grow beyond the 2000 used in this experiment. If we combine NT kernel adaptive filtering with EIPS-ITL estimate, e.g., $4$-th order Taylor polynomial or $D_e = 5$ used in NT-KMEE-GQ-TS(error), we obtain results comparable to the linear filter with SIG (LMEE-SIG) and substantially faster time than kernel SIG methods, as shown in the bottom plot of Fig. \ref{fig:KMEE}. On the other hand, the LMEE-SIG performed the worst in maximizing the IP (equivalent to minimizing the error entropy), as shown in the top plot of Fig. \ref{fig:KMEE}. The NT-KMEE with EIPS-ITL estimate converged to the maximum IP at the same iteration step as conventional kernel methods (nonlinear rate is due to the $1/N^2$ normalization of the double sum vs. the $1/N$ for SIG) but using significantly lower, constant CPU time.

\section{Conclusion}\label{Sec:Conclusion}
In this paper, we proposed a family of fast, scalable, and accurate estimators for information theoretic learning using explicit inner product spaces. ITL replaces conventional second-order statistics for information theory descriptors based on non-parametric estimator of R\'{e}nyi entropy. ITL is conceptually different from standard kernel methods as it is based on kernel density estimation. Although ITL kernels need not to satisfy Mercer's condition, positive definiteness is preferred due to numerical stability in computation. An RKHS for ITL defined on a space of probability density functions simplifies statistical inference for supervised or unsupervised learning. ITL criteria take into account the higher-order statistical behavior of the systems and signals as desired. However, this comes at an increased cost of complexity. By extending the no-trick kernel method to ITL using EIPS feature mapping with constant complexity for certain problems, information extraction from the signal is improved without compromising its scalability. We outlined several methods (deterministic, random, and hybrid) to construct EIPS feature mappings. We demonstrated the superior performance of EIPS-ITL estimators and combined NT-kernel adaptive filtering using EIPS-ITL cost functions through experiments.

In the future, we will extend the EIPS framework to more advanced ITL algorithms and cost functions, such as unsupervised learning and reinforcement learning.

\bibliographystyle{IEEEtran}
\bibliography{IEEEabrv,references}{}

 \newcommand{\noop}[1]{}
\begin{thebibliography}{10}
\providecommand{\url}[1]{#1}
\csname url@samestyle\endcsname
\providecommand{\newblock}{\relax}
\providecommand{\bibinfo}[2]{#2}
\providecommand{\BIBentrySTDinterwordspacing}{\spaceskip=0pt\relax}
\providecommand{\BIBentryALTinterwordstretchfactor}{4}
\providecommand{\BIBentryALTinterwordspacing}{\spaceskip=\fontdimen2\font plus
\BIBentryALTinterwordstretchfactor\fontdimen3\font minus
  \fontdimen4\font\relax}
\providecommand{\BIBforeignlanguage}[2]{{%
\expandafter\ifx\csname l@#1\endcsname\relax
\typeout{** WARNING: IEEEtran.bst: No hyphenation pattern has been}%
\typeout{** loaded for the language `#1'. Using the pattern for}%
\typeout{** the default language instead.}%
\else
\language=\csname l@#1\endcsname
\fi
#2}}
\providecommand{\BIBdecl}{\relax}
\BIBdecl

\bibitem{ITL}
J.~C. Pr{\'{\i}}ncipe, \emph{Information Theoretic Learning: Renyi's Entropy
  and Kernel Perspectives}.\hskip 1em plus 0.5em minus 0.4em\relax New York,
  NY, USA: Springer, 2010.

\bibitem{Parzen1962}
\BIBentryALTinterwordspacing
E.~Parzen, ``On estimation of a probability density function and mode,''
  \emph{Ann. Math. Statist.}, vol.~33, no.~3, pp. 1065--1076, 09 1962.
  [Online]. Available: \url{https://doi.org/10.1214/aoms/1177704472}
\BIBentrySTDinterwordspacing

\bibitem{Xu08}
{Jian-Wu Xu}, A.~R.~C. {Paiva}, I.~{Park}, and J.~C. {Principe}, ``A
  reproducing kernel hilbert space framework for information-theoretic
  learning,'' \emph{IEEE Transactions on Signal Processing}, vol.~56, no.~12,
  pp. 5891--5902, Dec 2008.

\bibitem{Liu10}
W.~Liu, J.~C. Pr{\'{\i}}ncipe, and S.~Haykin, \emph{Kernel Adaptive Filtering:
  A Comprehensive Introduction}.\hskip 1em plus 0.5em minus 0.4em\relax
  Hoboken, NJ, USA: Wiley, 2010.

\bibitem{KAARMA}
K.~Li and J.~C. Pr{\'{\i}}ncipe, ``The kernel adaptive
  autoregressive-moving-average algorithm,'' \emph{IEEE Trans. Neural Netw.
  Learn. Syst.}, vol.~27, no.~2, pp. 334--346, Feb. 2016.

\bibitem{Li2018}
\BIBentryALTinterwordspacing
------, ``Biologically-inspired spike-based automatic speech recognition of
  isolated digits over a reproducing kernel hilbert space,'' \emph{Frontiers in
  Neuroscience}, vol.~12, p. 194, 2018. [Online]. Available:
  \url{https://www.frontiersin.org/article/10.3389/fnins.2018.00194}
\BIBentrySTDinterwordspacing

\bibitem{Li2019functional}
\BIBentryALTinterwordspacing
K.~Li and J.~C. Principe, ``Functional bayesian filter,'' 2019. [Online].
  Available: \url{https://arxiv.org/abs/1911.10606}
\BIBentrySTDinterwordspacing

\bibitem{QKLMS}
B.~Chen, S.~Zhao, P.~Zhu, and J.~C. Pr{\'{\i}}ncipe, ``Quantized kernel least
  mean square algorithm,'' \emph{IEEE Trans. Neural Netw. Learn. Syst.},
  vol.~23, no.~1, pp. 22--32, 2012.

\bibitem{NICE}
K.~Li and J.~C. Príncipe, ``Transfer learning in adaptive filters: The nearest
  instance centroid-estimation kernel least-mean-square algorithm,'' \emph{IEEE
  Transactions on Signal Processing}, vol.~65, no.~24, pp. 6520--6535, Dec
  2017.

\bibitem{SNIPGOAL}
K.~{Li} and J.~C. {Príncipe}, ``Surprise-novelty information processing for
  gaussian online active learning (snip-goal),'' in \emph{2018 International
  Joint Conference on Neural Networks (IJCNN)}, July 2018, pp. 1--6.

\bibitem{SmolaSparseGreedy00}
\BIBentryALTinterwordspacing
A.~J. Smola and B.~Sch\"{o}kopf, ``Sparse greedy matrix approximation for
  machine learning,'' in \emph{Proceedings of the Seventeenth International
  Conference on Machine Learning}, ser. ICML '00.\hskip 1em plus 0.5em minus
  0.4em\relax San Francisco, CA, USA: Morgan Kaufmann Publishers Inc., 2000,
  pp. 911--918. [Online]. Available:
  \url{http://dl.acm.org/citation.cfm?id=645529.657980}
\BIBentrySTDinterwordspacing

\bibitem{WilliamsIDD2000}
\BIBentryALTinterwordspacing
C.~K.~I. Williams and M.~Seeger, ``The effect of the input density distribution
  on kernel-based classifiers,'' in \emph{Proceedings of the Seventeenth
  International Conference on Machine Learning}, ser. ICML '00.\hskip 1em plus
  0.5em minus 0.4em\relax San Francisco, CA, USA: Morgan Kaufmann Publishers
  Inc., 2000, pp. 1159--1166. [Online]. Available:
  \url{http://dl.acm.org/citation.cfm?id=645529.756511}
\BIBentrySTDinterwordspacing

\bibitem{Fine02}
\BIBentryALTinterwordspacing
S.~Fine and K.~Scheinberg, ``Efficient svm training using low-rank kernel
  representations,'' \emph{J. Mach. Learn. Res.}, vol.~2, pp. 243--264, Mar.
  2002. [Online]. Available:
  \url{http://dl.acm.org/citation.cfm?id=944790.944812}
\BIBentrySTDinterwordspacing

\bibitem{Seth09}
S.~{Seth} and J.~C. {Principe}, ``On speeding up computation in information
  theoretic learning,'' in \emph{2009 International Joint Conference on Neural
  Networks}, June 2009, pp. 2883--2887.

\bibitem{Li2019notrick}
\BIBentryALTinterwordspacing
K.~Li and J.~C. Principe, ``No-trick (treat) kernel adaptive filtering using
  deterministic features,'' 2019. [Online]. Available:
  \url{https://arxiv.org/abs/1912.04530}
\BIBentrySTDinterwordspacing

\bibitem{GREENGARD1987FMM}
L.~Greengard and V.~Rokhlin, ``A fast algorithm for particle simulations,''
  \emph{Journal of Computational Physics}, vol.~73, no.~2, pp. 325 -- 348,
  1987.

\bibitem{Greengard1991}
\BIBentryALTinterwordspacing
L.~Greengard and J.~Strain, ``The fast gauss transform,'' \emph{SIAM J. Sci.
  Stat. Comput.}, vol.~12, no.~1, pp. 79--94, Jan. 1991. [Online]. Available:
  \url{https://doi.org/10.1137/0912004}
\BIBentrySTDinterwordspacing

\bibitem{Yang2003}
{Yang}, {Duraiswami}, {Gumerov}, and {Davis}, ``Improved fast gauss transform
  and efficient kernel density estimation,'' in \emph{Proceedings Ninth IEEE
  International Conference on Computer Vision}, Oct 2003, pp. 664--671 vol.1.

\bibitem{Li13KAFCI}
K.~{Li}, B.~{Chen}, and J.~C. {Príncipe}, ``Kernel adaptive filtering with
  confidence intervals,'' in \emph{The 2013 International Joint Conference on
  Neural Networks (IJCNN)}, Aug 2013, pp. 1--6.

\bibitem{Dao2017}
T.~Dao, C.~D. Sa, and C.~R{\'e}, ``Gaussian quadrature for kernel features,''
  in \emph{Proceedings of the 31st International Conference on Neural
  Information Processing Systems}, ser. NIPS'17.\hskip 1em plus 0.5em minus
  0.4em\relax USA: Curran Associates Inc., 2017, pp. 6109--6119.

\bibitem{rahimi2007RFF}
A.~Rahimi and B.~Recht, ``Random features for large-scale kernel machines,'' in
  \emph{Proceedings of the 20th International Conference on Neural Information
  Processing Systems}, ser. NIPS'07.\hskip 1em plus 0.5em minus 0.4em\relax
  USA: Curran Associates Inc., 2007, pp. 1177--1184.

\bibitem{Williams00Nystrom}
\BIBentryALTinterwordspacing
C.~K.~I. Williams and M.~Seeger, ``Using the nystr\"{o}m method to speed up
  kernel machines,'' in \emph{Proceedings of the 13th International Conference
  on Neural Information Processing Systems}, ser. NIPS'00.\hskip 1em plus 0.5em
  minus 0.4em\relax Cambridge, MA, USA: MIT Press, 2000, pp. 661--667.
  [Online]. Available: \url{http://dl.acm.org/citation.cfm?id=3008751.3008847}
\BIBentrySTDinterwordspacing

\bibitem{Bach05ICD}
\BIBentryALTinterwordspacing
F.~R. Bach and M.~I. Jordan, ``Predictive low-rank decomposition for kernel
  methods,'' in \emph{Proceedings of the 22Nd International Conference on
  Machine Learning}, ser. ICML '05.\hskip 1em plus 0.5em minus 0.4em\relax New
  York, NY, USA: ACM, 2005, pp. 33--40. [Online]. Available:
  \url{http://doi.acm.org/10.1145/1102351.1102356}
\BIBentrySTDinterwordspacing

\bibitem{Bochner1959}
\BIBentryALTinterwordspacing
S.~Bochner, M.~Functions, S.~Integrals, H.~Analysis, M.~Tenenbaum, and
  H.~Pollard, \emph{Lectures on Fourier Integrals. (AM-42)}.\hskip 1em plus
  0.5em minus 0.4em\relax Princeton University Press, 1959. [Online].
  Available: \url{http://www.jstor.org/stable/j.ctt1b9s09r}
\BIBentrySTDinterwordspacing

\bibitem{Smolyak63}
{S. A. Smolyak}, ``Quadrature and interpolation formulas for tensor products of
  certain classes of functions,'' \emph{Dokl. Akad. Nauk SSSR}, vol. 148,
  no.~5, pp. 1042--1045, 1963.

\bibitem{Aronszajn1950}
N.~Aronszajn, ``Tehory of reproducing kernels,'' \emph{Trans. Amer. Math.
  Soc.}, vol.~68, pp. 337--404, 1950.

\bibitem{Parzen1959}
E.~Parzen, ``Statistical methods on time series by {H}ilbert space methods,''
  Applied Mathematics and Statistics Laboratory, Stanford, CA, Tech. Rep. 23,
  1959.

\bibitem{LiuCorrentropy2007}
W.~{Liu}, P.~P. {Pokharel}, and J.~C. {Principe}, ``Correntropy: Properties and
  applications in non-gaussian signal processing,'' \emph{IEEE Transactions on
  Signal Processing}, vol.~55, no.~11, pp. 5286--5298, Nov 2007.

\bibitem{Santamaria2006}
I.~Santamaria, P.~P. Pokharel, and J.~C. Pr{\'{\i}}ncipe, ``Generalized
  correlation funciton: definition, properties, and application to blind
  equalization,'' \emph{IEEE Trans. Signal Process.}, vol.~54, no.~6, pp.
  2187--2197, 2006.

\bibitem{KLMS}
W.~Liu, P.~Pokharel, and J.~C. Pr{\'{\i}}ncipe, ``The kernel least-mean-square
  algorithm,'' \emph{{IEEE} Trans. Signal Process.}, vol.~56, no.~2, pp.
  543--554, 2008.

\bibitem{KMCC}
S.~{Zhao}, B.~{Chen}, and J.~C. {Príncipe}, ``Kernel adaptive filtering with
  maximum correntropy criterion,'' in \emph{The 2011 International Joint
  Conference on Neural Networks}, July 2011, pp. 2012--2017.

\bibitem{HAN20072733}
\BIBentryALTinterwordspacing
S.~Han, S.~Rao, D.~Erdogmus, K.-H. Jeong, and J.~Principe, ``A minimum-error
  entropy criterion with self-adjusting step-size (mee-sas),'' \emph{Signal
  Processing}, vol.~87, no.~11, pp. 2733 -- 2745, 2007. [Online]. Available:
  \url{http://www.sciencedirect.com/science/article/pii/S0165168407001788}
\BIBentrySTDinterwordspacing

\bibitem{Erdogmus2002}
\BIBentryALTinterwordspacing
D.~Erdogmus and J.~C. Principe, ``Generalized information potential criterion
  for adaptive system training,'' \emph{Trans. Neur. Netw.}, vol.~13, no.~5,
  pp. 1035--1044, Sep. 2002. [Online]. Available:
  \url{https://doi.org/10.1109/TNN.2002.1031936}
\BIBentrySTDinterwordspacing

\bibitem{KMEE}
\BIBentryALTinterwordspacing
B.~Chen, Z.~Yuan, N.~Zheng, and J.~C. Pr\'{\i}ncipe, ``Kernel minimum error
  entropy algorithm,'' \emph{Neurocomput.}, vol. 121, pp. 160--169, Dec. 2013.
  [Online]. Available: \url{http://dx.doi.org/10.1016/j.neucom.2013.04.037}
\BIBentrySTDinterwordspacing

\bibitem{UCI_data}
\BIBentryALTinterwordspacing
``{UCI} machine learning repository.'' [Online]. Available:
  \url{http://archive.ics.uci.edu/ml}
\BIBentrySTDinterwordspacing

\bibitem{Mackey77}
M.~C. Mackey and L.~Glass, ``Oscillation and chaos in physiological control
  systems,'' \emph{Science}, vol. 197, no. 4300, pp. 287--289, Jul. 1977.

\end{thebibliography}
\begin{IEEEbiography}
[{\includegraphics[width=1in,height=1.25in,clip,keepaspectratio]{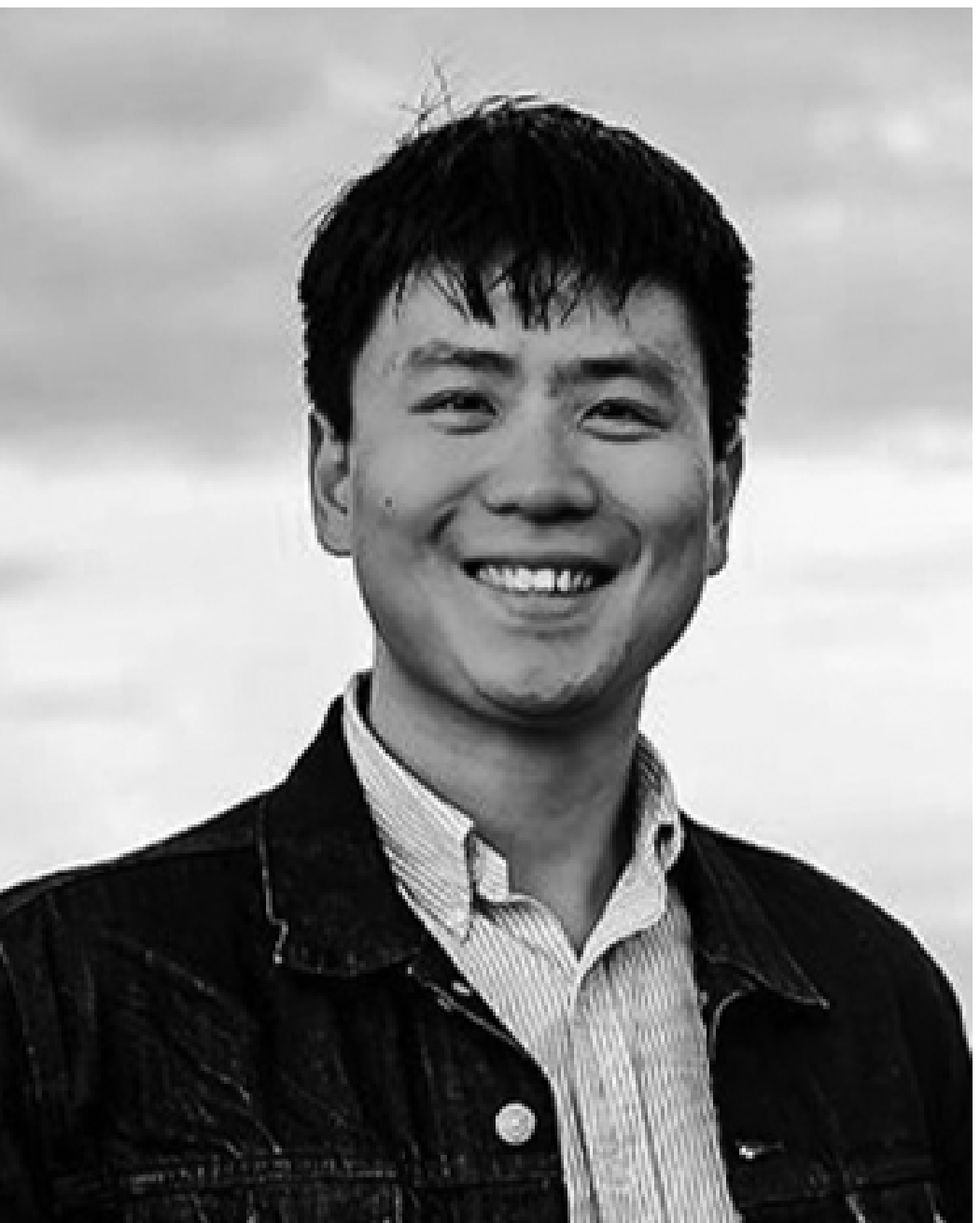}}]
{Kan Li} (S'08) received the B.A.Sc. degree in electrical engineering from the University of Toronto in 2007, the M.S. degree in electrical engineering from the University of Hawaii in 2010, and the Ph.D. degree in electrical engineering from the University of Florida in 2015.  He is currently a research scientist at the University of Florida. His research interests include machine learning and signal processing.
\end{IEEEbiography}
	
\vspace{-8 mm}
\begin{IEEEbiography}
[{\includegraphics[width=1in,height=1.25in,clip,keepaspectratio]{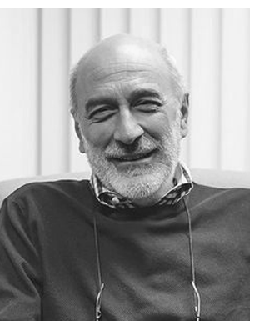}}]{Jos\'{e} C. Pr\'{i}ncipe}
(M'83-SM'90-F'00) is the BellSouth and Distinguished Professor of Electrical and Biomedical Engineering at the University of Florida, and the Founding Director of the Computational NeuroEngineering Laboratory (CNEL). His primary research interests
are in advanced signal processing with information theoretic criteria and adaptive models in reproducing kernel Hilbert spaces (RKHS), with application to brain-machine interfaces (BMIs). Dr.  Pr\'{i}ncipe is a Fellow of the IEEE, ABME, and AIBME. He is the past Editor in Chief of the IEEE Transactions on Biomedical Engineering, past Chair of the Technical Committee on Neural Networks of the IEEE Signal Processing Society, Past-President of the International Neural Network Society, and a recipient of the IEEE EMBS Career Award and the IEEE Neural Network Pioneer
		Award.
\end{IEEEbiography}	
\end{document}